\documentclass[runningheads]{llncs}
\DeclareUnicodeCharacter{2212}{-}

\usepackage{subcaption}

\usepackage{graphicx}
\usepackage{pdfpages}
\usepackage[colorlinks,linkcolor=blue]{hyperref}
\begin{document}

\title{BHSD: A 3D Multi-Class Brain Hemorrhage Segmentation Dataset} 



%
%
\author{Biao Wu\inst{1} \and
Yutong Xie\inst{1} \and
Zeyu Zhang\inst{1,3} \and 
Jinchao Ge\inst{1} \and 
Kaspar Yaxley\inst{2} \and 
Suzan Bahadir\inst{2} \and 
Qi Wu \inst{1} \and 
Yifan Liu \inst{1,4} \and 
Minh-Son To\inst{2} 
}

\institute{ Australia Institute of Machine Learning, University of Adelaide \and
Flinders Health and Medical Research Institute, Flinders University \and
Australian National University \and
ETH Zürich 
}



%

%
\maketitle              

\begin{abstract}

Intracranial hemorrhage (ICH) is a pathological condition characterized by bleeding inside the skull or brain, which can be attributed to various factors. 
Identifying, localizing and quantifying ICH has important clinical implications, in a bleed-dependent manner. 
While deep learning techniques are widely used in medical image segmentation and have been applied to the ICH segmentation task, existing public ICH datasets do not support the multi-class segmentation problem. 
To address this, we develop the Brain Hemorrhage Segmentation Dataset (BHSD), which provides a 3D multi-class ICH dataset containing 192 volumes with pixel-level annotations and 2200 volumes with slice-level annotations across five categories of ICH. 
To demonstrate the utility of the dataset, we formulate a series of supervised and semi-supervised ICH segmentation tasks. 
We provide experimental results with state-of-the-art models as reference benchmarks for further model developments and evaluations on this dataset. The dataset and checkpoint is available at \url{https://github.com/White65534/BHSD}.

\end{abstract}

\section{Introduction}

Intracranial hemorrhage (ICH) refers to bleeding that occurs inside the skull or brain. There are various types, based on the anatomical relation of the bleeding with the brain and its surrounding membranes. These types include extradural hemorrhage (EDH), subdural hemorrhage (SDH), subarachnoid hemorrhage (SAH), intraparenchymal hemorrhage (IPH), and intraventricular hemorrhage (IVH). The causes of ICH are diverse~\cite{howard2013risk,gronbaek2008liver,mccarron1999cerebral}, encompassing factors such as trauma, vascular malformations, tumors, hypertension, and venous thrombosis. The detection and characterization of ICH are typically done using non-contrast CT scans, which can reveal the type and distribution of bleeding. Accurate localization, quantification, and classification of hemorrhages are crucial as they have significant clinical implications, allowing clinicians to estimate severity, predict outcomes, and monitor progress~\cite{steiner2014european,hemphill2015guidelines,auer1989endoscopic,fisher06aSAH}. Treatment options may also vary depending on the quantity and type of bleed. 

The rapid development of deep learning and large-scale labeled dataset has accelerated the automation of medical image segmentation~\cite{wang2021deep,lee2018practical}. Despite the importance of delineating different classes of hemorrhage, many existing public ICH datasets focus either on hemorrhage classification~\cite{hssayeni2020computed} or single-class segmentation~\cite{li2023state}, specifically foreground versus background. Thus, it is essential to develop tools that can accurately segment different classes of hemorrhage rather than solely perform foreground versus background segmentation. Unfortunately, the lack of public datasets with multi-class, pixel-level annotations hinders the development of class-specific segmentation techniques.

The purpose of this work is to augment a large, public ICH dataset\cite{flanders2020construction} to produce a 3D, multi-class ICH dataset with pixel-level hemorrhage annotations, hereafter referred to as the brain hemorrhage segmentation dataset (BHSD). 
Our approach leverages the existing high-quality slice-level annotations performed by neuroradiologists and subsequently relabels a subset of CT scans with multi-class pixel-level annotations. 
We demonstrate the utility of this dataset by performing a series of experiments and providing benchmarks on supervised and semi-supervised segmentation tasks. 

\section{Multi-Class Brain Hemorrhage Segmentation Dataset}

\subsection{Brain hemorrhage datasets}



\begin{figure}[t]
\centering
     \begin{subfigure}[b]{0.48\textwidth}
         \includegraphics[width=\textwidth]{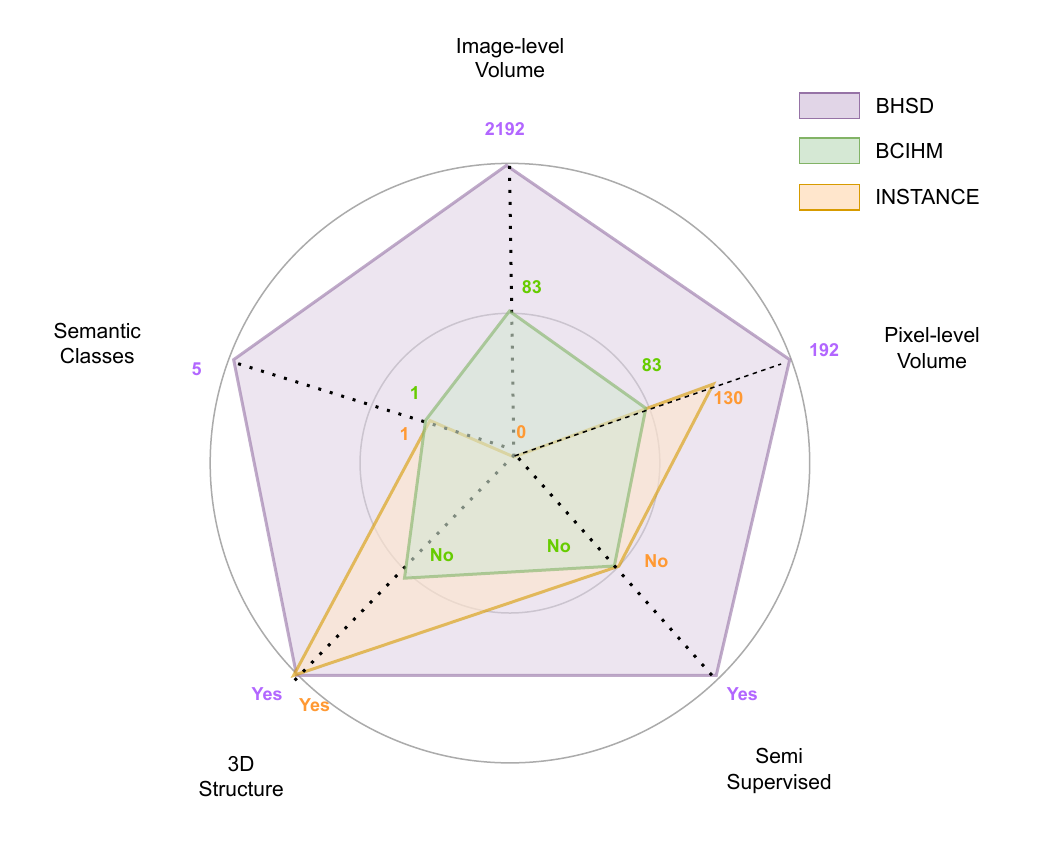}
         \caption[]{Comparison with other datasets.} 
         \label{fig:BHSDcomparison}
     \end{subfigure}
     \qquad
     \begin{subfigure}[b]{0.4\textwidth}
         \includegraphics[width=\textwidth]{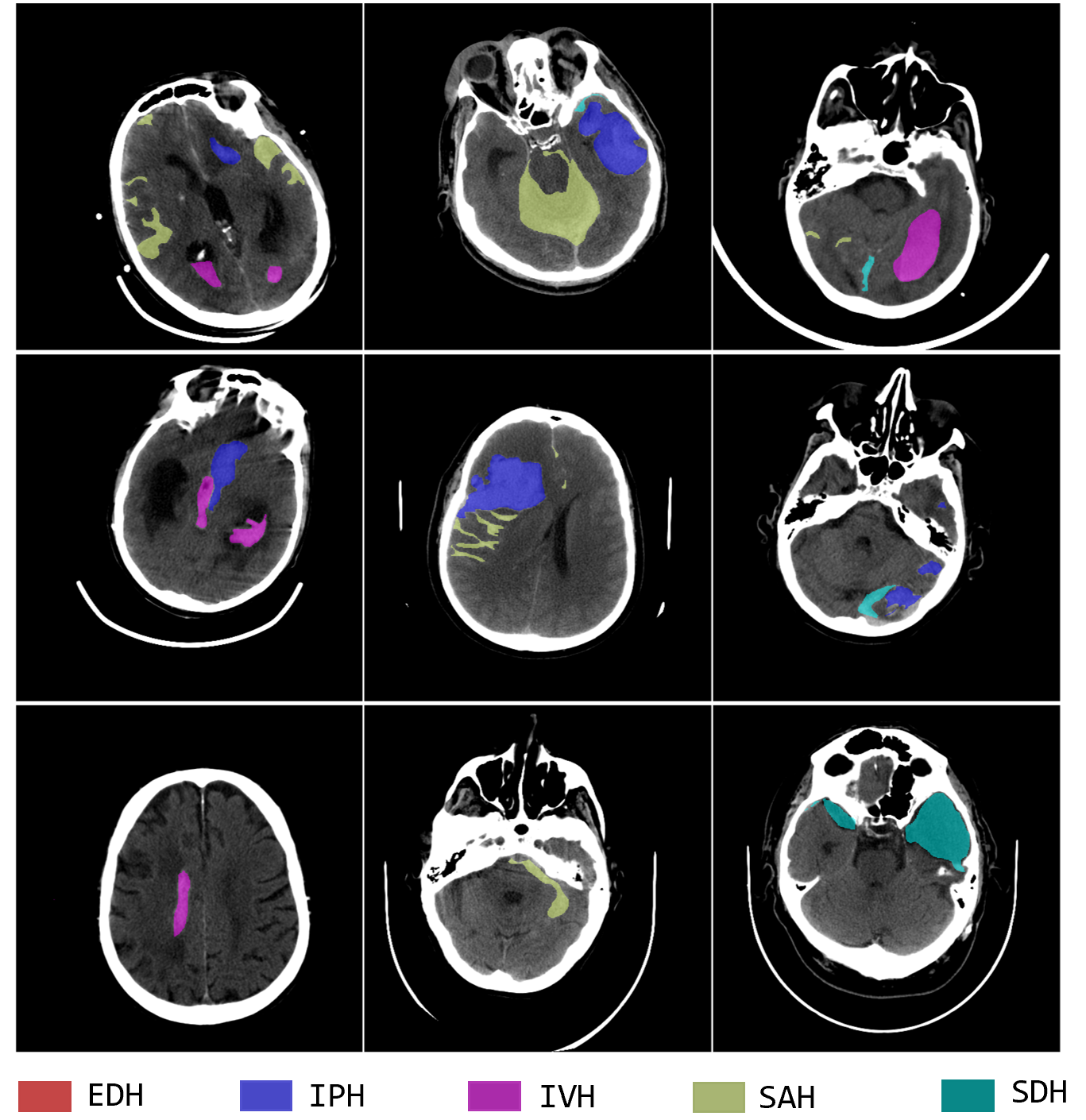}
         \caption[]{Example annotations.} 
         \label{fig:BHSDexamples}
     \end{subfigure}

    \caption{Overview of the BHSD. (a) Comparison of the BCIHM dataset, INSTANCE dataset and BHSD in terms of data characteristics. The BHSD provides more data, and more annotation information at different levels. (b) Representative examples of the diverse hemorrhage annotations provided in the BHSD. The different colors correspond to different hemorrhage classes, as indicated by the legend.} 
    \label{fig:BHSDcomposition}
\end{figure}

In this section, we describe existing, public brain hemorrhage datasets. 

The BCIHM dataset consists of 82 non-contrast CT scans of patients with traumatic brain injury \cite{hssayeni2020computed}. The dataset is provided in NIfTI format. Each slice of the scans was reviewed by two radiologists who recorded hemorrhage types if hemorrhage occurred or if a fracture occurred. Hemorrhage regions in each slice were also segmented, however, the annotations only support a foreground versus background segmentation. 
In contrast, the INSTANCE dataset, introduced as part of a MICCAI 2022 Challenge, consists of 200 volumes, of which 130 have foreground and background segmentation labels \cite{li2023state}. The volumes in this dataset are also provided in NIfTI format, but again all bleed types are combined into a single foreground class.

The CQ500 dataset consists of 491 CT scans with 193,317 slices in DICOM format \cite{chilamkurthy2018development}. The scans have been read by three radiologists, and the annotations provided indicate, at the scan level, the presence, type and location of hemorrhage. While this dataset does not support segmentation tasks, an augmentation of this dataset, BHX, provides bounding box annotations for five types of hemorrhage \cite{BHX2020}. Specifically, BHX contains 39,668 bounding boxes in 23,409 images. Another key brain hemorrhage dataset was published by the Radiological Society of North America (RSNA) \cite{flanders2020construction}. This dataset is a public collection of 874,035 CT head images in DICOM format from a mixed patient cohort with and without ICH. The dataset is multi-institutional and multi-national and includes slice-level expert annotations from neuroradiologists about the presence and type of bleed. This dataset was used for the RSNA 2019 Machine Learning Challenge for detecting brain hemorrhages, \textit{i.e.,} a classification, not segmentation problem.

\subsection{Reconstruction and annotation pipeline of the BHSD}

The BHSD is a high-quality medical imaging dataset comprising 2192 high-resolution 3D CT scans of the brain, each containing between 24 to 40 slices of 512 × 512 pixels in size (Fig. \ref{fig:BHSDcomparison}). 
The original RSNA dataset was provided as a collection of randomly sorted slices in DICOM format with slice-level annotations. 
Important contextual information in adjacent slices may be lost in single slices, hence the first step was to reconstitute 3D head scans. 
Since the anonymized patient identifiers were provided and the DICOM files retained geometric/positional data, the original 3D head scans could be reconstructed and converted to NIfTI format \cite{larobina2014medical}. 
The slice-level hemorrhage labels provided with the original RSNA dataset were mapped to the corresponding slices in the reconstructed head scans. 
A subset of 192 scans with one or more of five categories of ICH, namely EDH, IPH, IVH, SAH, and/or SDH, was subsequently selected for further annotation. 

Pixel-level annotations were performed by three medical imaging experts in two stages. 
Hemorrhages on individual head scans were independently segmented using ITK-SNAP \cite{yushkevich2017itk} by two trained medical imaging experts and radiology residents, both with over one year of experience reading CT head scans, using the original image-level hemorrhage annotations as a guide. These annotations were then reviewed by a board-certified radiologist with over 5 years post-fellowship experience, ensuring the quality of the annotations. 
To supplement the 192 volumes with pixel-level annotations, we also collected  corresponding image-level annotations from the RSNA dataset and provide an additional 2000 3D CT scans with slice-level annotations, including scans with no bleed. The composition of the BHSD is shown in Fig. \ref{fig:BHSDcomposition}. 

By covering both image-level and pixel-level annotations, the BHSD  allows a more comprehensive interrogation of brain hemorrhage imaging, and as we show, enables the development of more varied deep learning methods for ICH segmentation.

\begin{figure}[t]
\centering
     \begin{subfigure}[b]{0.39\textwidth}
         \includegraphics[width=\textwidth]{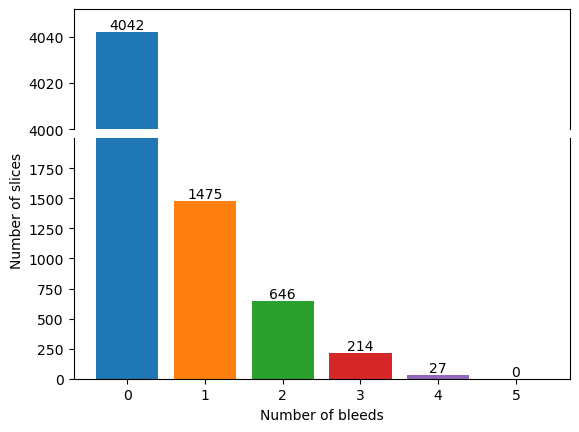}
         \caption[]{Number of bleeds (slice level)} 
         \label{fig:Afull}
     \end{subfigure}
     \qquad
     \begin{subfigure}[b]{0.39\textwidth}
         \includegraphics[width=\textwidth]{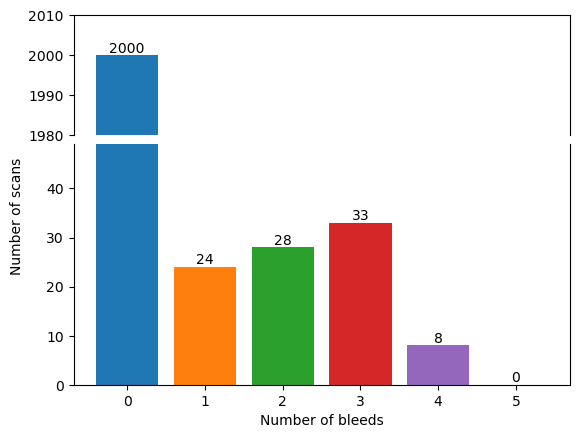}
         \caption[]{Number of bleeds (scan level)} 
         \label{fig:Bfull}
     \end{subfigure}

     \vskip\baselineskip
     \begin{subfigure}[b]{0.39\textwidth}
         \includegraphics[width=\textwidth]{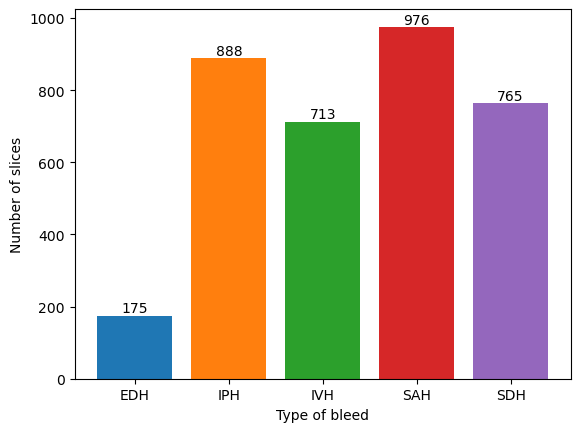}
         \caption[]{Type of bleed (slice level)} 
         \label{fig:Azoomed}
     \end{subfigure}
     \qquad
     \begin{subfigure}[b]{0.39\textwidth}
         \centering
         \includegraphics[width=\textwidth]{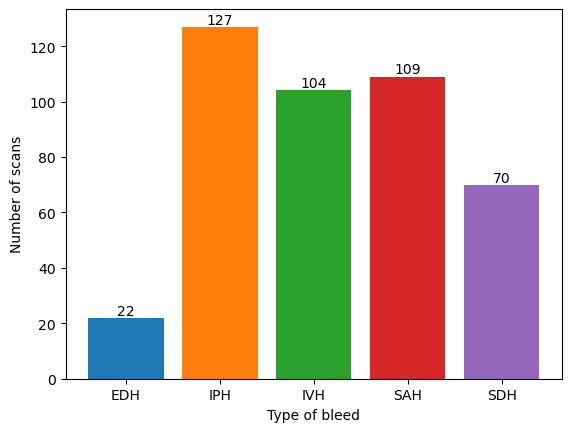}
         \caption[]{Type of bleed (scan level)} 
         \label{fig:Bzoomed}
     \end{subfigure}
    \caption{Summary composition of the BHSD, at the slice and scan levels, by number and type of bleed.} 
    \label{fig:BHSDcomposition}
\end{figure}





    


    



\subsection{Segmentation applications using the BHSD}

In this section, we describe two segmentation applications designed to leverage the proposed BHSD.



\textbf{Supervised segmentation.} 
Supervised multi-class segmentation refers to the classification of all individual pixels in an image into distinct classes, using segmentation mask annotations for supervision. 
For ICH, multi-class segmentation is clinically more significant while technically more challenging compared with foreground and background segmentation. 
Multi-class segmentation can more accurately identify and segment different types of ICH, which is important for diagnosis, treatment planning and prognostication. 
However, multi-class  segmentation also has many challenges since ICH come in different shapes, sizes, and densities, and multiple different types of bleeding may occur simultaneously. 
This task may therefore require more data than the foreground/background problem.
We evaluate the performance of supervised segmentation methods under different conditions using the BHSD. 

\textbf{Semi-supervised with pixel-labeled and unlabeled data.} 
Acquiring labeled medical imaging data can be a costly and challenging process, whereas unlabeled data is usually more obtainable. 
Annotating medical images also requires specialized domain expertise, which can pose a significant barrier to the widespread development of deep learning methods for clinical practice. 
Semi-supervised learning (SSL) addresses this challenge by using a small amount of labeled data and a large amount of unlabeled data for model training. 
Using the BHSD, we simulate the SSL scenario by discarding the image-level annotations to build an unlabeled dataset. 
Our approach combines data with pixel-level annotations and unlabeled data to evaluate the performance of SSL methods. 

\section{Experiments and Benchmarking Methods}

To demonstrate the utility of the BHSD, we perform a series of segmentation experiments under different conditions and provide benchmarks for future model evaluations using this dataset.



\subsubsection{Evaluation metrics}
Segmentation performance was evaluated using the Dice similarity coefficient (DSC) \cite{dice1945measures}, which compares the similarity between the predicted and true segmentation.

\subsubsection{Implementation details}
All experiments were performed on a single A6000 GPU.
Following nnUnet~\cite{isensee2021nnu}, we first truncated the HU values of each scan using the range of [−40,120], and then normalized the truncated voxel values by subtracting 40 and dividing by 80. We randomly cropped sub-volumes of size [32,128,128] from CT scans as the input.
Other parameters in different models retain the official default settings. In the BHSD, the 192 volumes were divided evenly into training and testing sets. The sets were balanced in terms of the number and types of bleeds, each containing 96 volumes. Furthermore, by verifying the original patient identifiers, no patient was contained in both sets.

\subsection{Supervised 3D segmentation }

In Experiment 1, we conducted a comprehensive evaluation of state-of-the-art 3D semantic segmentation models using the BHSD dataset. We evaluate five 3D models with SOTA backbones designed for supervised semantic segmentation, namely UNETR \cite{hatamizadeh2022unetr}, Swin UNETR \cite{hatamizadeh2022swin,liu2021swin}, CoTr \cite{xie2021cotr}, nnFormer \cite{zhou2021nnformer,vaswani2017attention}, and nnUNet \cite{isensee2021nnu} (Table \ref{table:3Dbenchmark}). 



\begin{table}[h]
\centering
\caption{Benchmark 3D supervised segmentation performance using the BHSD. }\label{table:3Dbenchmark}
\begin{tabular}{l|l|l|l|l|l|l}
\hline
 &  EDH & IPH & IVH &  SAH &  SDH & Mean \\
\hline 
\hline
UNETR & 1.64 & 28.28 & 22.08 & 4.36 & 3.63 & 11.99      \\ 
Swin UNETR &  2.53 & 34.18 & 29.28 & 10.07 & 8.43 & 16.89 \\
CoTr   & 1.63 & 48.62 & 53.55 & 17.88 & 15.44 & 27.43   \\ 
nnFormer & 0.00 & 69.75 & 25.78 & 25.94 & 10.31 & 29.19 \\
nnUnet3D & 4.81 & 54.12 & 51.48 & 21.57 & 15.23 & 29.44  \\ 
\hline
\end{tabular}
\end{table}

The results indicated that nnUnet achieved superior performance compared to other models with an average Dice of 29.44. However, it is important to acknowledge the limitation of the dataset, specifically the low occurrence of epidural hematoma \textit{i.e.,} EDH class. Consequently, the segmentation performance for this class was considerably inferior to other classes. Further refinement and adaptation are necessary to enhance the segmentation performance for the EDH class, considering its limited representation in the dataset. Confusion matrix analysis may also allow identification of imbalances and biases in model predictions (Supplementary Fig. 1).


\subsection{Incorporation of scans with no hemorrhage}

In Experiment 2, we sought to enhance the model's performance through a gradual augmentation of the training set with negative samples within a supervised experimental setup. 
This also allowed us to address the issue of false positives. 
The model achieved optimal performance with 200 negative samples (Table \ref{table:negative}), both in terms of hemorrhage segmentation performance and suppression of false positives. 



\begin{table}
\centering
\caption{Incorporation of scans with (B) and without bleeding (NB). The false positive (FP) rate is also indicated. }
\label{tab2}
\begin{tabular}{l|l|l|l|l|l|l|l}
\hline
Heading level &  EDH & IPH & IVH &  SAH &  SDH & Mean & FP  \\
\hline 
 
96B         &   4.81   &   54.12    & 51.48      & 21.57      & 15.23     & 29.44 & 1.41 \\ 
96B + 200NB &   9.77   &   56.90    &  58.53     &   29.98    &   21.73   & 35.38 & 0.84 \\
96B + 400NB &   4.46   &   39.45    &  39.90     &   15.43    &   9.30    & 21.71 & 1.30 \\
96B + 600NB &   3.98   &    36.24   &   26.25    &   6.41    &    6.59   & 15.90 & 1.85 \\
96B + 800NB &  2.14    &  28.21     &   29.57    &  4.47     &   2.78    & 13.43 & 2.41 \\
\hline
\end{tabular}
\label{table:negative}
\end{table}



This outcome suggests the presence of a balance point, where an insufficient number of negative samples hampers the model's ability to effectively learn the discriminating features, leading to inadequate suppression of false positives, while excessive negative samples may cause the model to overly focus on them, resulting in difficulties distinguishing between target and non-target categories (Supplementary Fig. 2). 
It is worth noting that selection of the optimal number of negative samples requires a comprehensive consideration of the model's recognition and generalization abilities, as well as the characteristics and requirements of the dataset. 
Future research can explore combinations of varying sample quantities to gain a deeper understanding and achieve more precise performance optimization.

\subsection{Single class, multiple models versus multiple class, single model} 

The utilization of existing brain hemorrhage data for multi-class semantic segmentation necessitates training a single-class detection model and merging the prediction outcomes. To further investigate the advantages of the BHSD dataset compared to existing datasets, in Experiment 3 we performed separate training iterations on BHSD, focusing on one category at a time. We repeated this process five times, resulting in distinct models capable of recognizing individual categories. Subsequently, we combined the inference results from these five models. 
Through comparative analysis, we observed that the multi-class semantic method surpassed its predecessor in the fusion result (Table \ref{table:multiclass}). However, it is important to note that the application of model fusion for single-category segmentation yielded considerable challenges, such as extensive overlapping and conflicting prediction outcomes (Supplementary Fig. 3). 
This finding underscores the ineffectiveness of the fusion approach for multi-class semantic segmentation.


\begin{table}
\centering
\caption{Multiple single class models versus single multi-class model. }
\label{table:multiclass}
\begin{tabular}{c|c|c|c|c|c|c}
\hline
Heading level &  EDH & IPH & IVH &  SAH &  SDH & Mean  \\
\hline 
1class-EDH & 5.15 &     -  &     -  &   -    &   -    & -  \\ 
1class-IPH &   -   &   37.55 &    -   &    -   &   -    & -  \\
1class-IVH &   -   &    -   &  23.60 &    -   &   -    & -  \\
1class-SAH &   -   &    -   &   -    &  14.50 &   -    & -  \\
1class-SDH &   -   &    -   &   -    &    -   & 17.59  &    \\
5class-Merge  &   1.25   &    3.42   &   2.57    &    2.43   &  2.16     & 19.68   \\
5class-Single & 4.81 & 54.12 & 51.48 & 21.57 & 15.23 & 29.44  \\ 
\hline
\end{tabular}
\end{table}







\vspace{-5mm}
\subsection{Semi-supervised segmentation} 

Experiment 4 was conducted with the aim of investigating the utilization of unlabelled data from clinical settings to enhance the performance of the dataset. Through the implementation of semi-supervised experiments, we sought to assess the potential for improvement. The unlabelled data introduced in this experiment were randomly sampled, with no available information regarding the health status or presence of hemorrhage. 
To evaluate this setting, four methods were applied to the BHSD, namely mean teacher~\cite{tarvainen2017mean}, cross pseudo supervision (CPS)~\cite{chen2021semi}, entropy minimization~\cite{vu2019advent}, and interpolation consistency~\cite{verma2022interpolation}. 
In these experiments, the training set of 96 volumes was supplemented by 500 unlabeled data. The same test set of 96 volumes was retained. For this task, we also merged hemorrhage labels to a single foreground mask. 

The experimental findings revealed that the model's performance could indeed be further enhanced by employing an appropriate semi-supervised approach (Table \ref{table:SSL}). This observation highlights the efficacy of incorporating unlabelled data in clinical settings. Moreover, the merging of all categories served to accentuate the extent of performance improvement achieved through this approach.
While there was mixed performance, we find that CPS improves on supervised learning and achieves 49.50\% DSC in the binary segmentation task. 
The CPS method~\cite{chen2021semi} uses pseudo-labels generated by one model to train another model, and then using this new model to generate new pseudo-labels and so forth, iterating this process to improve the accuracy of the pseudo-labelled data and improve the performance of the model segmentation.

\begin{table}[ht]
\centering
\caption{Semi-supervised performance based on nnUNet. To highlight the advantages of semi-supervision, we merged all the hemorrhage categories and report foreground and background semantic segmentation results.}\label{table:SSL}
\begin{tabular}{l|c|cccccccc}
\hline
Method  & unlabeled samples   & Dice \\
\hline
\hline
SupOnly                    & 0       & 45.10 $\pm$0.21  \\  
Entropy Minimization         & 500     & 36.91 $\pm$0.16 \\ 
Mean Teacher                  & 500     & 44.63 $\pm$0.18 \\  
Interpolation Consistency      & 500     & 45.38 $\pm$0.33 \\ 
Cross Pseudo Supervision        & 500     & 49.50 $\pm$0.19 \\ 
\hline 
\end{tabular}
\end{table}


\section{Conclusion}

We describe a 3D CT head dataset, the BHSD, for intracranial hemorrhage segmentation. This dataset includes a diverse mix of head scans with pixel-level and slice-level annotations, as well as scans with and without hemorrhage. To qualitatively and quantitatively scrutinize the characteristics of the BHSD, we compare popular SOTA models and diverse training techniques and draw three key insights from our benchmarking experiments. 
Firstly, the BHSD can significantly enhance the performance of SOTA models for multi-class segmentation of ICH. Multi-class segmentation significantly outperforms the fusion of single-class segmentation models. 
Secondly, incorporation of scans without hemorrhage can enhance segmentation performance. However, the right balance needs to be found.
Third, the BHSD improves model performance using a semi-supervised approach even when the volumes are not annotated at the pixel-level.  
Hence, the BHSD is a valuable dataset for ICH segmentation models, providing an opportunity to study how segmentation tasks can make better use of unlabeled and weakly labeled data.
This will in turn facilitate the development and validation of computer-aided diagnostic tools for clinical practice. 




%
%
%

\end{document}